\documentclass[11pt,letterpaper]{article}
\usepackage{acl2012}
\usepackage{times}
\usepackage{latexsym}
\usepackage{xcolor}
\usepackage{amsmath, graphicx, amsfonts, amssymb}
\usepackage{algorithmicx}
\usepackage{algorithm}
\usepackage[noend]{algpseudocode}
\usepackage{breqn}
\makeatletter
\newcommand{\@BIBLABEL}{\@emptybiblabel}
\newcommand{\@emptybiblabel}[1]{}
\makeatother
\usepackage{hyperref}
\usepackage{cleveref}
\usepackage{subcaption}
\usepackage{multirow, array} 
\hypersetup{
    citecolor=black,
    filecolor=black,
    linkcolor=black,
    urlcolor=black
}

 \usepackage[disable]{todonotes} 

 \newcommand{\comment}[1]{}  
\newcommand*{\helvet}{\textsl}
\newcommand*{\update}{\textcolor{black}} 

\newcommand*{\updateB}{\textcolor{black}} 

\title{An Unsupervised Method for Uncovering Morphological Chains}

\author{Karthik Narasimhan,  Regina Barzilay and  Tommi Jaakkola\\
	    CSAIL, Massachusetts Institute of Technology \\	    
	    {\tt \{karthikn, regina, tommi\}@csail.mit.edu}\\
	}

\date{}

\begin{document}
\maketitle
\begin{abstract}

Most state-of-the-art systems today produce morphological analysis based only on orthographic patterns. In contrast, we propose a model for unsupervised morphological analysis that integrates orthographic and semantic views of words. We model word formation in terms of morphological chains, from base words to the observed words, breaking the chains into parent-child relations. We use log-linear models with morpheme and word-level features to predict possible parents, including their modifications, for each word. The limited set of candidate parents for each word render contrastive estimation feasible. Our model consistently \updateB{matches or} outperforms five state-of-the-art systems on Arabic, English and Turkish.\footnote{\update{Code is available at \url{https://github.com/karthikncode/MorphoChain}.}}
\end{abstract}
\section{Introduction}

Morphologically related words exhibit connections at multiple levels, ranging from orthographical patterns to semantic proximity. For instance, the words \helvet{playing} and \helvet{played} share the same stem, but also carry similar meaning. Ideally, all these complementary sources of information would be taken into account when learning morphological structures.   

Most state-of-the-art unsupervised approaches to morphological analysis are built primarily around
orthographic patterns in morphologically-related words~\cite{Goldwater:2004:PBL:1622153.1622158,Creutz:2007:UMM:1187415.1187418,Snyder08unsupervisedmultilingual,Poon:2009:UMS:1620754.1620785}. In these approaches, words are commonly modeled as concatenations of morphemes. This morpheme-centric view is well-suited for uncovering distributional properties of stems and affixes. But it is not well-equipped to capture semantic relatedness at the word level. 

In contrast, earlier approaches that capture semantic similarity in morphological variants operate solely at
 the word level~\cite{Schone:2000:KIM:1117601.1117615,journals/corr/cs-CL-0205006}. Given two candidate words, the proximity is assessed using standard word-distributional measures such as mutual information. However, the fact that these models do not model morphemes directly greatly limits their performance. 

In this paper, we propose a model to integrate orthographic and semantic views. Our goal is to build a chain of derivations for a current word from its base form. For instance, given a word \helvet{playfully}, the corresponding chain is \helvet{play} $\rightarrow$ \helvet{playful} $\rightarrow$ \helvet{playfully}. The word \helvet{play} is a base form of this derivation as it cannot be reduced any further. Individual derivations are obtained by adding a morpheme (ex. \helvet{-ful}) to a parent word (ex. \helvet{play}). This addition may be implemented via a simple concatenation, or it may involve transformations. At every step of the chain, the model aims to find a parent-child pair (\update{ex. \helvet{play-playful}}) such that the parent also constitutes a valid entry in the lexicon. This allows the model
to directly compare the semantic similarity of the parent-child pair, while also considering the orthographic properties of the morphemic combination.

We model each step of a morphological chain by means of a log-linear model that enables us to incorporate a wide range of features. At the semantic level, we consider the relatedness between two words using 
the corresponding vector embeddings. At the orthographic level, features capture whether the words in the chain actually occur in the corpus, how affixes are reused, as well as how the words are altered during the addition of morphemes. We use Contrastive Estimation~\cite{Smith:2005:CET:1219840.1219884} to efficiently learn this model in an unsupervised manner. Specifically, we require that each word has greater support among its bounded set of candidate parents than an artificially constructed neighboring word would.

  
We evaluate our model on datasets in three languages: Arabic, English and Turkish. We compare our performance 
against five state-of-the-art unsupervised systems: Morfessor Baseline~\cite{virpioja2013report}, Morfessor CatMAP~\cite{morfessor}, AGMorph~\cite{journals/tacl/SirtsG13}, the Lee Segmenter~\cite{Lee:2011:MSC:2018936.2018937,stallard2012unsupervised} and the system of~\newcite{Poon:2009:UMS:1620754.1620785}. \update{Our model consistently equals or outperforms these systems across the three languages. For instance, on English, we obtain an 8.5\% gain in F-measure over Morfessor. Our experiments also demonstrate the value of semantic information. While the contribution varies from 3\% on Turkish to 11\% on the English dataset, it nevertheless improves performance across all the languages.} 

\section{Related Work}

Currently, top performing unsupervised morphological analyzers are based on the orthographic properties of sub-word units~\cite{morfessor,Creutz:2007:UMM:1187415.1187418,Poon:2009:UMS:1620754.1620785,journals/tacl/SirtsG13}. Adding semantic information to these systems is not an easy task, as they operate at the level of individual morphemes, rather than morphologically related words.  
 
 The value of semantic information has been demonstrated in earlier work on morphological analysis. \newcite{Schone:2000:KIM:1117601.1117615} employ an LSA-based similarity measure to identify morphological variants from a list of orthographically close word pairs. The filtered pairs are then used to identify stems and affixes.  Based on similar intuition, \newcite{journals/corr/cs-CL-0205006} design a method that integrates these sources of information, captured as two word pair lists, ranked based on edit distance and mutual information. These lists are subsequently combined using a deterministic weighting function. 
 
In both of these algorithms, orthographic relatedness is based on simple deterministic rules. Therefore,
semantic relatedness plays an essential role in the success of these methods. However, these algorithms do not capture distributional properties of morphemes that are critical to the success of current state-of-the-art algorithms. In contrast, we utilize a single statistical framework that seamlessly combines both sources of information. Moreover, it allows us to incorporate a wide range of additional features.

Our work also relates to the log-linear model for morphological segmentation developed by \newcite{Poon:2009:UMS:1620754.1620785}. They propose a joint model over all words (observations) and their segmentations (hidden), using morphemes and their contexts (character n-grams) for the features.  Since the space of all possible \emph{segmentation sets} is huge, learning and inference are quite involved. They use techniques like Contrastive Estimation, sampling and simulated annealing. 
In contrast, our formulation does not result in such a large search space. For each word, the number of parent candidates is bounded by its length multiplied by the number of possible transformations. Therefore,
Contrastive Estimation can be implemented via enumeration, and does not require sampling.   Moreover, 
operating at the level of words (rather than morphemes) enables us to incorporate semantic and word-level features.

\update{Most recently, work by \newcite{journals/tacl/SirtsG13} uses Adaptor Grammars for minimally supervised segmentation. By defining a morphological grammar consisting of zero or more prefixes, stems and suffixes,  they induce segmentations over words in both unsupervised and semi-supervised settings. While their model (AGMorph) builds up a word by combining morphemes in the form of a parse tree, we operate at the word level and build up the final word via intermediate words in the chain. 
}

\update{In other related work, \newcite{dreyer2011discovering} tackle the problem of recovering morphological paradigms and inflectional principles. They use a Bayesian generative model with a log-linear framework, using expressive features, over pairs of strings.  Their work, however, handles a different task from ours and requires a small amount of annotated data to seed the model.}

\update{ In this work, we make use of semantic information to help morphological analysis. \newcite{Lee:2011:MSC:2018936.2018937} present a model that takes advantage of syntactic context to perform better morphological segmentation. \updateB{\newcite{stallard2012unsupervised} improve on this approach using the technique of Maximum Marginal decoding to reduce noise}. Their best system considers entire sentences, while our approach (and the morphological analyzers described above) operates at the vocabulary level without regarding sentence context. Hence, though their work is not directly comparable to ours, it presents an interesting orthogonal view to the problem. 
}

\section{Model}
\subsection{Definitions and Framework}

We use \emph{morphological chains} to model words in the language. A morphological chain is a short sequence of words that starts from the base word and ends up in a morphological variant. Each node in the chain is, by assumption, a valid word. We refer to the word that is morphologically changed as the \emph{parent} word \update{and its morphological variant as the \emph{child} word}. A word that does not have any morphological parent is a \emph{base word} (e.g., words like \helvet{play}, \helvet{chat}, \helvet{run}).\footnote{We distinguish base words from morphological \emph{roots} which do not strictly speaking have to be valid words in the language.} 

Words in a chain (other than the base word) are created from their parents by adding morphemes (prefixes, suffixes, or other words). For example, a morphological chain that ends up in the word \helvet{internationally} could be \helvet{nation} $\rightarrow$ \helvet{national} $\rightarrow$ \helvet{international} $\rightarrow$ \helvet{internationally}. The base word for this chain is \helvet{nation}. Note that the same word can belong to multiple morphological chains. 
 For example, the word \helvet{national} appears also as part of another chain that ends up in \helvet{nationalize}. These chains are treated separately but with shared statistical support for the common parts. For this reason, our model breaks morphological chains into possible parent-child relations such as (\helvet{nation}, \helvet{national}).

\begin{table}
\centering
\begin{tabular}{| c | c |} \hline
\textbf{Segment} & \textbf{Cosine Similarity} \\ \hline
p & 0.095 \\
pl & -0.037 \\
pla & -0.041 \\
play & \textbf{0.580} \\
playe & 0.000 \\
player & 1.000 \\ \hline
\end{tabular}
\caption{Cosine similarities between word vectors of various segments of the word \helvet{player} and the vector of \helvet{player}.}
\label{cosine}
\end{table}

We use a log-linear model for predicting parent-child pairs.  A log-linear model allows an easy, efficient way of incorporating several different features pertaining to parent-child relations. In our case, we leverage both orthographic and semantic patterns to encode representative features. 

A log-linear model consists of a set of features represented by a feature vector $\phi : \mathcal{W} \times \mathcal{Z} \rightarrow  \mathbb{R}^d $ and a corresponding weight vector $ \theta \in \mathbb{R}^d$. Here, $\mathcal{W}$ is a set of words and $\mathcal{Z}$ is the set of  \emph{candidates} for words in $\mathcal{W}$, that includes the parents as well as their types. Specifically, a \emph{candidate} is a (\emph{parent, type}) pair, where the \emph{type} variable keeps track of the type of morphological change (or the lack thereof if there is no parent) as we go from the parent to the child. In our experiments, $\mathcal{Z}$ is obtained by collecting together all sub-words created by splitting observed words in $\mathcal{W}$ at all different points. For instance, if we take the word \helvet{cars}, the candidates obtained by splitting would include (\helvet{car}, \textit{Suffix}), (\helvet{ca}, \textit{Suffix}), (\helvet{c}, \textit{Suffix}), (\helvet{ars}, \textit{Prefix}), (\helvet{rs}, \textit{Prefix}) and (\helvet{s}, \textit{Prefix}).

Note that the parent may undergo changes as it is joined with the affix and thus, there are more choices for the parent than just the ones obtained by splitting. Hence, to the set of candidates, we also add modified sub-words where transformations include character repetition (\helvet{plan} $\rightarrow$ \helvet{planning}), deletion (\helvet{decide} $\rightarrow$ \helvet{deciding}) or replacement (\helvet{carry
} $\rightarrow$ \helvet{carried}).\footnote{We found that restricting the set of parents to sub-words that are at least half the length of the original word helped improve the performance of the system.} Following the above example for the word \helvet{cars}, we get candidates like (\helvet{cat}, \textit{Modify}) and (\helvet{cart}, \textit{Delete}). Each word also has a stop candidate \mbox{(-, \emph{Stop})}, which is equivalent to considering it as a base word with no parent.

Let us define the probability of a particular word-candidate pair ($w \in \mathcal{W}, z \in \mathcal{Z}$) as $P(w, z) \propto e^{\theta \cdot \phi(w,z)} $.
The conditional probability of a candidate $z$ given a word $w$ is then
 \begin{equation}
 	 P(z | w) =  \frac{e^{\theta \cdot \phi(w,z)}}{\sum_{z' \in C(w)} e^{\theta \cdot \phi(w,z')}},\;\;z \in C(w) \nonumber
 \end{equation}
where   $C(w) \subset \mathcal{Z}$ refers to the set of possible candidates (parents and their types) for the word $w \in \mathcal{W}$.

In order to generate a possible ancestral chain for a word, we recursively predict parents until the word is predicted to be a base word.  In our model, these choices are included in the set of candidates for the specific word, and their likelihood is controlled by the type related features. Details of these and other features are given in section \ref{sec:features}.


\subsection{Features}
\label{sec:features}
 \begin{table*}
\centering
\begin{tabular}{| c | c | c | c | c |} \hline
\textbf{Feature type} & \textbf{Word (w)} & \textbf{Candidate (p,t)} & \textbf{Feature} & \textbf{Value} \\ \hline 
Cosine & painter & (paint, Suffix) & $\vec{w} \cdot \vec{p}$  & 0.58 \\ \hline
Affix & painter & (paint, Suffix) & \textit{suffix}=er  & 1\\ \hline
Affix Correlation & walking & (walk, Suffix) & \textit{AffixCorr(ing, ed)} & 1 \\ \hline
Wordlist & painter & (paint, Suffix) & \textit{WordFreq} &  8.73\\ 
 &  & & \textit{OutOfVocab} &  0\\ \hline
Transformations & planning & (plan, Repeat) & \textit{type=Repeat $\times$ chars=(n,-)} & 1 \\
 & deciding & (decide, Delete) & \textit{type=Delete $\times$ chars=(e,-)} & 1 \\
 & carried &  (carry, Modify) & \textit{type=Modify $\times$ chars=(y,i)} & 1 \\ \hline
Stop & painter & (-, Stop) & \textit{begin=pa} & 1 \\
 & & & \textit{end=er} & 1 \\
 & & & 0.5 $<$ \textit{MaxCos $<$ 0.6} & 1 \\
  & & & \textit{length}=7 & 1 \\ \hline
\end{tabular}
\caption{Example of various types of features used in the model. $\vec{w}$ and $\vec{p}$ are the word vectors for the word and parent, respectively.}
\label{features}
\end{table*}

This section provides an overview of the features used in our model. The features are defined for a given word $w$, parent $p$ and type $t$ (recall that a candidate $z \in \mathcal{Z}$ is the pair $(p, t)$). For computing some of these features, we use an unannotated list of words with frequencies (details in section~\ref{sec:setup}). Table \ref{features} provides a summary of the features.

\paragraph{Semantic Similarity} We hypothesize that morphologically related words exhibit semantic similarity. To this end, we introduce a feature that measures cosine similarity between the word vectors of the word~($\vec{w}$) and the parent~($\vec{p}$).  These word vectors are learned from co-occurrence patterns from a large corpus\footnote{For strings which do not have a vector learnt from the corpus, we set the cosine value to be -0.5.} (see section~\ref{sec:setup} for details). 

To validate this measure, we computed the cosine similarity between words and their morphological parents from the CELEX2 database~\cite{celex2}. On average, the resulting word-parent similarity score is 0.351, compared to 0.116 for randomly chosen word-parent combinations.\footnote{The cosine values range from around -0.1 to 0.7 usually.}

\paragraph{Affixes} A distinctive feature of affixes is their frequent occurrence in multiple words. To capture this pattern, we automatically generate a list of frequently occurring candidate affixes. These candidates are collected by considering the string difference between a word and its parent candidates which appear in the word list. For example, for the word \helvet{paints}, possible suffixes include \helvet{-s} derived from the parent \helvet{paint}, \helvet{-ts} from the parent \helvet{pain} and \helvet{-ints} from the word \helvet{pa}.
Similarly, we compile a list of potential prefixes. These two lists are sorted by their frequency and thresholded. For each affix in the lists, we have a corresponding indicator variable. For unseen affixes, we use an UNK (unknown) indicator.

These automatically constructed lists act as a proxy for the gold affixes. In English, choosing the top 100 suffixes in this manner gives us 43 correct suffixes (compared against gold suffixes). Table~\ref{suff-examples} gives some examples of suffixes generated this way.

\begin{table}
\centering
\resizebox{\columnwidth}{!}{%
\begin{tabular}{| c | c  |} \hline 
\textbf{Language} & \textbf{Top suffixes} \\ \hline
English & -s, \mbox{-'s}, -d, -ed, -ing, -', -s', -ly, -er, -e \\
 Turkish & -n, -i, -lar, -dir, -a, -den, -de, -in, -leri, -ler \\
 Arabic & -p, -A, -F, -y, -t, -AF, -h, -hA, -yp, -At\\ \hline
\end{tabular}
}
\caption{Top ten suffixes in automatically produced suffix lists.}
\label{suff-examples}
\end{table}

\paragraph{Affix Correlation} While the previous feature considers one affix assignment at a time, there is a  known correlation between affixes attached to the same stem. For instance, in English, verbs that can be modified by the suffix \helvet{-ing}, can also take the related suffix \helvet{-ed}.  Therefore, we introduce a feature that measures, whether for a given affix and parent, we also observe in the wordlist the same parent modified by its related affix. For example, for the pair \helvet{(walking, walk)}, the feature instance \mbox{\emph{AffixCorr(ing, ed)}} is set to 1, because the word \helvet{walked} is in the WordList.

To construct pairs of related affixes, we compute the correlation between pairs in auto-generated affix list described previously.  This correlation is proportional to the number of stems the two affixes share. For English, examples of such pairs include \helvet{(inter-, re-), (under-, over-), (-ly, -s)}, and \helvet{(-er, -ing)}.

\paragraph{Presence in Wordlist} We want to bias the model to select parents that constitute valid words.\footnote{This is not an absolute requirement in the model.} Moreover, we would like to take into account the frequency of the parent words. We encode this information as the logarithm of their word counts in the wordlist (\emph{WordFreq}). For parents not in the wordlist, we set a binary \emph{OutOfVocab} feature to 1.  

\paragraph{Transformations} We also support transformations to enable non-concatenative morphology. Even in English, which is mostly concatenative, such transformations are frequent. We consider three kinds of transformations previously considered in the literature~\cite{Goldwater:2004:PBL:1622153.1622158}:
\begin{itemize}
\item repetition of the last character in the parent (ex.~\helvet{plan} $\rightarrow$ \helvet{planning})
\item deletion of the last character in the parent (ex.~\helvet{decide} $\rightarrow$ \helvet{deciding})
\item modification of the last character of the parent (ex.~\helvet{carry} $\rightarrow$ \helvet{carried})
\end{itemize}
 
We add features that are the cartesian product of the type of transformation and the character(s) involved. For instance, for the parent-child pair (\helvet{believe}, \helvet{believing}), the feature \textit{type=Delete $\times$ chars=(e,-)}  will be activated, while the rest of the transformational features will be 0. 

\paragraph{Stop Condition} Finally, we introduce features that aim to identify base words which do not have a parent. The features include the length of the word, and the starting and ending character unigrams and bigrams. In addition, we include a feature that records the highest cosine similarity between the word and any of its candidate parents. This feature will help, for example, to identify \helvet{paint} as a base word, instead of choosing \helvet{pain} as its parent.

\subsection{Learning}

 We learn the log-linear model in an unsupervised manner without explicit feedback about correct morphological segmentations. We assume that we have an unannotated wordlist $D$ for this purpose. A typical approach to learning such a model would be to maximize the likelihood of all the observed words in $D$ over the space of all strings constructible in the alphabet, $\Sigma^*$, by marginalizing over the hidden candidates.\footnote{We also tried maximizing instead of marginalizing, but the model gets stuck in one of the numerous local optima.} In other words, we could use the EM-algorithm to maximize
 \begin{dmath}
 	 L(\theta;D) = \prod_{w^* \in D} P(w^*) = \prod_{w^* \in D} \sum_{z \in C(w^*)} P(w^*,z)
	 	 = \prod_{w^* \in D} \left[ \frac{\sum_{z \in C(w^*)} e^{\theta \cdot \phi(w^*,z)}}{\sum_{w \in \Sigma^*} \sum_{z \in C(w)} e^{\theta \cdot \phi(w,z)}} \right] 
	 \label{em}
 \end{dmath}
 However, maximizing $L(\theta;D)$ is problematic since approximate methods would be needed to sum over $\Sigma^*$ in order to calculate the normalization term in \eqref{em}. Moreover, we would like to encourage the model to emphasize relevant parent-child pairs\footnote{In other words, assign higher probability mass.} out of a smaller set of alternatives rather than those pertaining to all the words. 
 
We employ Contrastive Estimation~\cite{Smith:2005:CET:1219840.1219884} and replace the normalization term by a sum over the \emph{neighbors} of each word. For each word in the language, we create neighboring strings in two sets. For the first set, we \emph{transpose} a single pair of adjacent characters of the word. We perform this transposition over the first $k$ or the last $k$ characters of the word.\footnote{The performance increases with increasing $k$ until $k=5$, after which no gains were observed.} For the second set, we transpose two pairs of characters simultaneously -- one from the first $k$ characters and one from the last $k$.

The combined set of artificially constructed words represents the events that we wish to move probability mass away from in favor of the actually observed words. 
The neighbors facilitate the learning of good weights for the affix features by providing the required contrast (at both ends of the words) to the actual words in the vocabulary. A remaining concern is that the model may not distinguish any arbitrary substring from a good suffix/prefix. For example, \mbox{\helvet{-ng}} appears in all the words that end with \helvet{-ing}, and could be considered a valid suffix. We include other features to help make this distinction. Specifically, we include features such as word vector similarity and the presence of the parent in the observed wordlist. For example, in the word \helvet{painting}, the parent candidate \helvet{paint} is likely to occur and also has a high cosine similarity with \helvet{painting} in terms of their word vectors. In contrast, \helvet{painti} does not. 
 
Given the list of words and their neighborhoods, we define the contrastive likelihood as follows:
\begin{dmath}
 	 L_{CE}(\theta,D) = \prod_{w^* \in D} \left[ \frac{\sum_{z \in C(w^*)} e^{\theta \cdot \phi(w^*,z)}}{\sum_{w \in N(w^*)} \sum_{z \in C(w)} e^{\theta \cdot \phi(w,z)}} \right] 
	 \label{opt}
 \end{dmath}
where $N(w^*)$ is the neighborhood of $w^*$.  This likelihood is much easier to evaluate and optimize. 
  
After adding in a standard regularization term, we maximize the following log likelihood objective:
 \begin{dmath}
 	 \sum_{w^* \in D} \left[ \log \sum_{z \in C(w^*)} e^{\theta \cdot \phi(w^*,z)} - \log \sum_{w \in N(w^*)} \sum_{z \in C(w)} e^{\theta \cdot \phi(w,z)} \right] - \lambda ||\theta||^2 
	 \label{opt2}
 \end{dmath}
The corresponding gradient can be derived as:
  \begin{dmath}
 	  \frac{\partial L_{CE}(\theta;D)} {\partial \theta_j} = \sum_{w^* \in D} \left[\frac{\sum_{z \in C(w^*)} \phi_j(w^*,z) \cdot e^{\theta \cdot \phi(w^*,z)}} {\sum_{z \in C(w^*)} e^{\theta \cdot \phi(w^*,z)}} - \frac{ \sum_{w \in N(w^*)} \sum_{z \in C(w)} \phi_j(w,z) \cdot e^{\theta \cdot \phi(w,z)}} { \sum_{w \in N(w^*)} \sum_{z \in C(w)} e^{\theta \cdot \phi(w,z)}} \right]  - 2 \lambda \theta_j
	  \label{gradient}
 \end{dmath}
 
We use LBFGS-B~\cite{lbfgs} to optimize $L_{CE}(\theta;D)$ with gradients given above. 

 \subsection{Prediction}
 Given a test word, we predict a morphological chain in a greedy step by step fashion. In each step, we use the learnt weights to predict the best parent for the current word (from the set of candidates), or choose to stop and declare the current word as a base word if the stop case has the highest score. Once we have the chain, we can derive a morphological segmentation by inserting a segmentation point (into the test word) appropriately for each edge in the chain. 
 
Algorithms  \ref{alg:predict} and \ref{alg:chain} provide details on the prediction procedure.
In both algorithms,  \emph{type} refers to the type of modification (or lack of) that the parent undergoes: Prefix/Suffix addition, types of transformation like repetition, deletion, modification, or the Stop case.

  \begin{algorithm}
\caption{Procedure to predict a parent for a word}
\label{alg:predict}
\begin{algorithmic}[1]
\Procedure{predict}{word}
\State $candidates \gets \Call{candidates}{word}$
\State $bestScore \gets 0$
\State $bestCand \gets (-, STOP)$
\For {$ cand \in candidates$}
	\State $features \gets \Call{features}{word, cand}$
	\State $score \gets \Call{modelScore}{features}$
	\If {$ score > bestScore $}
		\State $bestScore \gets score$
		\State $bestCand \gets cand$
	\EndIf
\EndFor
\State \Return $bestCand$
\EndProcedure
\end{algorithmic}
\end{algorithm}

\begin{algorithm}[h]
\caption{Procedure to predict a morphological chain}
\label{alg:chain}
\begin{algorithmic}[1]
\Procedure{getChain}{word}
\State $candidate \gets \Call{predict}{word}$
\State $parent, type \gets candidate$
\If {$type = STOP$} 
	\Return $[(word, \emph{STOP})]$
\EndIf
\State \Return \Call{getChain}{parent}+$[(parent, type)]$
\EndProcedure
\end{algorithmic}

\end{algorithm}

\section{Experimental Setup}
\label{sec:setup} 
\paragraph{Data}

We run experiments on three different languages: English, Turkish and Arabic. For each language, we utilize corpora for training, testing and learning word vectors. The training data consists of an unannotated wordlist with frequency information, while the test data is a set of gold morphological segmentations. For the word vectors, we train the word2vec tool~\cite{word2vec} on large text corpora and obtain 200-dimensional vectors for all three languages. Table~\ref{data_stats} provides information about each dataset.

 \begin{table}[!t]
\centering
\begin{tabular}{| c | c | c | c |} \hline
\textbf{Lang} & \textbf{Train} & \textbf{Test} & \textbf{WordVec} \\
 & \textbf{(\# words)} & \textbf{(\# words)} &  \textbf{(\# words)}\\ \hline
English & MC-10 & MC-05:10  & Wikipedia\\
& (878K) & (2218) &  (129M)\\ \hline
Turkish & MC-10 & MC-05:10 & BOUN  \\
  &  (617K) & (2534) & (361M)\\ \hline
Arabic & Gigaword & ATB & Gigaword\\
 &  (3.83M) & (119K) &  (1.22G)\\ \hline
 \end{tabular}
\caption[blah]{Data corpora and statistics. MC-10 = MorphoChallenge 2010\footnotemark, MC-05:10 = MorphoChallenges 2005-10 (aggregated), BOUN = BOUN corpus~{\cite{sak2008turkish}}, Gigaword = Arabic Gigaword corpus~{\cite{gigaword-arabic}}, ATB = Arabic Treebank~{\cite{treebank-arabic}} }
\label{data_stats}
\end{table}
\footnotetext{http://research.ics.aalto.fi/events/morphochallenge/}

 \paragraph{Evaluation measure}
We test our model on the task of morphological segmentation.
We evaluate performance on individual segmentation points, which is standard for this task~\cite{journals/tal/VirpiojaTSKK11}. We compare predicted segmentations against the gold test data for each language and report overall Precision, Recall and F-1 scores calculated across all segmentation points in the data. 
As is common in unsupervised segmentation~\cite{Poon:2009:UMS:1620754.1620785,journals/tacl/SirtsG13}, we included the test words (without their segmentations) with the training words during parameter learning.

\paragraph{Baselines}
 We  compare our model with five other systems: Morfessor Baseline (Morf-Base), Morfessor CatMap (Morf-Cat), AGMorph, the Lee Segmenter and the system of \newcite{Poon:2009:UMS:1620754.1620785}.  Morfessor has achieved excellent performance on the MorphoChallenge dataset, and is widely used for performing unsupervised morphological analysis on various languages, even in fairly recent work~\cite{Luong-etal:conll13:morpho}. In our experiments, we employ two variants of the system because their relative performance varies across languages. We use publicly available implementations of these variants~\cite{virpioja2013report,morfessor}. We perform several runs with various parameters, and choose the run with the best performance on each language. 
   
 \update{We evaluate AGMorph by directly obtaining the posterior grammars from the authors.\footnote{The grammars were trained using data we provided to them.} We show results for the Compounding grammar, which we find has the best average performance over the languages.} \updateB{The Lee Segmenter~\cite{Lee:2011:MSC:2018936.2018937}, improved upon by using Maximum Marginal decoding in \newcite{stallard2012unsupervised}, has achieved excellent performance on the Arabic (ATB) dataset. We perform comparison experiments with the model 2 (M2) of the segmenter, which employs latent POS tags, and does not require sentence context which is not available for other languages in the dataset. We obtained the code for the system, and run it on our English and Turkish datasets.\footnote{We report numbers on Arabic directly from their paper.}} We do not have access to an implementation of Poon et al's system; hence, we directly report scores from their paper on the ATB dataset and test our model on the same data.


\section{Results}
\label{sec:results}
\todo[inline]{check results after finalized}

Table~\ref{results} details the performance of the various models on the segmentation task. We can see that our method outperforms both variants of Morfessor, with an absolute gain of 8.5\%, 5.1\% and 5\% in F-score on English, Turkish and Arabic, respectively. 
 On Arabic, we obtain a 2.2\% absolute improvement over Poon et al.'s model. AGMorph doesn't segment better than Morfessor on English and Arabic but does very well on Turkish (60.9\% F1 compared to our model's 61.2\%). This could be due to the fact that the Compounding grammar is well suited to the agglutinative morphology in Turkish \updateB{and hence provides more gains than for English and Arabic. The Lee Segmenter (M2) performs the best on Arabic (82\% F1), but lags behind on English and Turkish. This result is consistent with the fact that the system was optimized for Arabic.}

 \begin{table}[!t]
\centering
\resizebox{\columnwidth}{!}{%
\begin{tabular}{| c | c | c | c | c |} \hline
\textbf{Lang} & \textbf{Method} & \textbf{Prec} & \textbf{Recall} & \textbf{F-1} \\ \hline
\multirow{7}{*}{English}  & \textit{Morf-Base} & 0.740 & 0.623 & 0.677 \\
  & \textit{Morf-Cat} & 0.673 & 0.587 & 0.627\\
  & \textit{AGMorph} & 0.696 & 0.604 & 0.647\\
   & \updateB{\textit{Lee (M2)}} & 0.825 & 0.525 & 0.642 \\
& Model -C & 0.555 & 0.792 & 0.653 \\ 
& Model -T & 0.831 & 0.664 & 0.738 \\ 
& Model -A & 0.810 & 0.713 & 0.758 \\ 
 & Full model & 0.807 & 0.722 & \textbf{0.762} \\ \hline
\multirow{7}{*}{Turkish}  & \textit{Morf-Base} & 0.827 & 0.362 & 0.504 \\
  & \textit{Morf-Cat} & 0.522 & 0.607 & 0.561\\
    & \textit{AGMorph} & 0.878 & 0.466 & 0.609 \\
     & \updateB{\textit{Lee (M2)}} & 0.787 & 0.355 & 0.489 \\
 & Model -C & 0.516 & 0.652 & 0.576 \\ 
 & Model -T & 0.665 & 0.521 & 0.584 \\ 
 & Model -A & 0.668 & 0.543 & 0.599 \\ 
 & Full model & 0.743 & 0.520 & \textbf{0.612} \\ \hline
 \multirow{8}{*}{Arabic}  & \textit{Morf-Base} & 0.807 & 0.204 & 0.326  \\
  & \textit{Morf-Cat} & 0.774 & 0.726 & 0.749 \\
    & \textit{AGMorph} & 0.672 &  0.761 & 0.713\\
 & \textit{Poon et al.} & 0.885 & 0.692 & 0.777 \\
 & \updateB{\textit{Lee (M2)}} & - & - & \textbf{0.820} \\
 & Model -C & 0.626 & 0.912 & 0.743 \\ 
 & Model -T & 0.774 & 0.807 & 0.790 \\ 
 & Model -A & 0.775 & 0.808 & 0.791 \\ 
 & Full model & 0.770 & 0.831 & 0.799 \\ \hline
\end{tabular}
}%
\caption{Results on unsupervised morphological segmentation; scores are calculated across all segmentation points in the test data. Baselines are in italics. -C=without cosine features, -T=without transformation features, -A=without affix correlation features. Numbers on Arabic for Poon et al. and Lee (M2) are reported directly from their papers.}
\label{results}
\end{table}

The table also demonstrates the importance of the added semantic information in our model. For all three languages, having the features that utilize cosine similarity provides a significant boost in performance. We also see that the transformation features and affix correlation features play a role in improving the results, although a less important one.

\begin{figure}[!t]
\centering
\resizebox{\columnwidth}{!}{%
\includegraphics[]{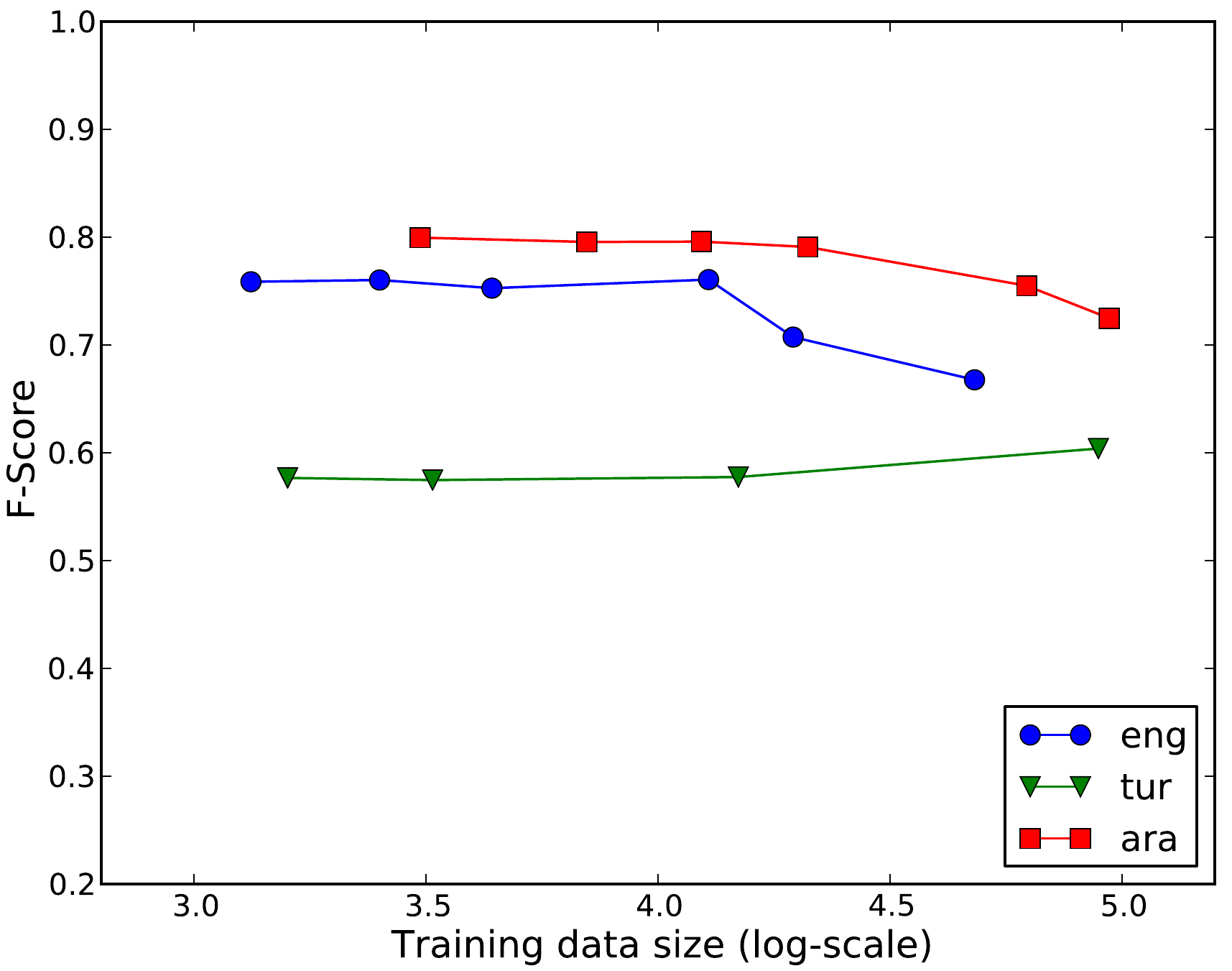}
}
\caption{Model performance vs data size obtained by frequency thresholding}
\label{fig:perf-vs-size}

\end{figure}

Next, we study the effect of data quality on the prediction of the algorithm. A training set often contains
misspellings, abbreviations and truncated words. Thresholding based on frequency is commonly used to reduce 
this noise. Figure~\ref{fig:perf-vs-size} shows the performance of the algorithm as a function of the data size obtained at various degrees of thresholding. We note that the performance of the model on all three languages remains quite stable from about 1000 to 10000 training words, after which the deviations are more apparent. The plot also demonstrates that the model works well even with a small amount of quality data ($\approx$3000 most frequent words).

 \paragraph{\update{Error analysis}}  
 
   \begin{table*}[!t]
\centering
\resizebox{\textwidth}{!}{%
\begin{tabular}{| c || c | c || c | c | c |} \hline
\multirow{2}{*}{\textbf{Language}} & \multicolumn{2}{ c|| }{\textbf{Correct Segmentations}} & \multicolumn{3}{ c| }{\textbf{Incorrect Segmentations}}\\ \cline{2-6}
 & \textbf{Word} & \textbf{Segmentation} & \textbf{Word} & \textbf{Predicted} & \textbf{Correct} \\ \hline
 \multirow{5}{*}{English} & salvoes  & salvo-es & contempt & con-tempt & contempt \\ 
 &  negotiations & negotiat-ion-s & sterilizing & steriliz-ing & steril-iz-ing \\
  & telephotograph & tele-photo-graph & desolating & desolating &  desolat-ing\\
   & unequivocally & un-equivocal-ly & storerooms & storeroom-s & store-room-s \\
    & carsickness's & car-sick-ness-'s & tattlers & tattler-s & tattl-er-s \\ \hline
\multirow{5}{*}{Turkish} & moderni  & modern-i & mektupla\c{s}malar & mektupla\c{s}ma-lar & mektup-la\c{s}-ma-lar\\
&  teknolojideki &  teknoloji-de-ki & gelecektiniz & gelecek-tiniz & gel-ecek-ti-niz \\
& buras{\i}yd{\i} & bura-s{\i}-yd{\i} & aynalardan & ayna-lar-da-n & ayna-lar-dan \\
& \c{c}izgisine & \c{c}iz-gi-si-ne & uyudu\u{g}unuzu & uyudu-\u{g}u-nuzu & uyu-du\u{g}-unuz-u \\
& de\u{g}i\c{s}iklikte & de\u{g}i\c{s}ik-lik-te & dirse\u{g}e & dirse\u{g}e & dirse\u{g}-e \\ \hline
 \multirow{5}{*}{Arabic}  & sy\$Ark & s-y-\$Ark & wryfAldw & w-ry-fAldw & w-ryfAldw \\
 & nyqwsyA & nyqwsyA & bHlwlhA & b-Hlwl-h-A & b-Hlwl-hA \\
 &  AlmTrwHp & Al-mTrwH-p & jnwby & jnwb-y & jnwby \\
 & ytEAmlwA & y-tEAml-wA & wbAyrn & w-bAyr-n & w-bAyrn \\
 & lAtnZr & lA-t-nZr & rknyp & rknyp &  rkny-p \\ \hline
\end{tabular}
}
\caption{Examples of correct and incorrect segmentations produced by our model on the three languages. Correct segmentations are taken directly from gold MorphoChallenge data.}
\label{error-examples}
\end{table*}

 \update{We look at a random subset of 50 incorrectly segmented words\footnote{Words with at least one segmentation point incorrect} in the model's output for each language. Table~\ref{errors} gives a breakup of errors in all 3 languages due to over or under-segmentation. Table~\ref{error-examples} provides examples of correct and incorrect segmentations predicted by our model.}
 
   \begin{table}[!t]
\centering
\begin{tabular}{| c | c | c |} \hline
\textbf{Lang} & \textbf{Over-segment} & \textbf{Under-segment}\\ \hline
English & 10\% & 86\% \\
Turkish  & 12\% & 78 \%\\
 Arabic  & 60\% &  40\% \\ \hline
\end{tabular}
\caption{Types of errors in analysis of 50 randomly sampled incorrect segmentations for each language. The remaining errors are due to incorrect placement  of segmentation points.}
\label{errors}
\end{table}

In English, most errors are due to under-segmentation of a word. We find that around 60\% of errors are due to roots that undergo transformations while morphing into a variant (see \cref{error-examples} for examples). 
 Errors in Turkish are also mostly due to under-segmentation. On further investigation, we find that most such errors (58\% of the 78\%) are due to parent words either not in vocabulary or having a very low word count ($\leq 10$). In contrast, we observe a majority of over-segmentation errors in Arabic (60\%). \updateB{This is likely because of Arabic having more single character affixes than the other languages. We find that 56\% of errors in Arabic involve a single-character affix, which is much higher than the 24.6\% that involve a two-letter affix. In contrast, 25\% of errors in English are due to single character affixes -- around the same number as the 24\% of errors due to two-letter affixes.}
 
Since our model is an unsupervised one, we make several simplifying assumptions to keep the candidate set size manageable for learning. For instance, we do not explicitly model infixes, since we select parent candidates by only modifying the ends of a word.
  Also, the root-template morphology of Arabic, a Semitic language, presents a complexity we do not directly handle. For instance, words in Arabic can be formed using specific patterns (known as \emph{binyanim}) (ex.  \helvet{nZr} $\rightarrow$ \helvet{yntZr}).  
 However, on going through the errors, we find that only 14\%  are due to these binyanim patterns not being captured.\footnote{This might be due to the fact that the gold segmentations also do not capture such patterns. For example, the gold segmentation for \helvet{yntZrwn} is given as \helvet{y-ntZr-wn}, even though \helvet{ntZr} is not a valid root.}  Adding in transformation rules to capture these types of language-specific patterns can help increase both chain and segmentation accuracy. 
 
 \paragraph{Analysis of learned distributions}
 To investigate how decisive the learnt model is, we examine the final probability distribution $P(z|w)$ of parent candidates for the words in the  English wordlist. We observe that the probability of the best candidate ($max_{z} P(z|w)$), averaged over all words,  is 0.77. We also find that the average entropy of the distributions is 0.65, which is quite low considering that the average number of candidates is 10.76 per word, which would result in a max possible entropy of around 2.37 if the distributions were uniform. This demonstrates that the model tends to prefer a single parent for every word,\footnote{Note that the candidate probability distribution may have more than a single peak in some cases.} which is exactly the behavior we want.

\begin{figure}[!t]
\centering
\begin{minipage}[b]{\linewidth}
\includegraphics[width=\linewidth]{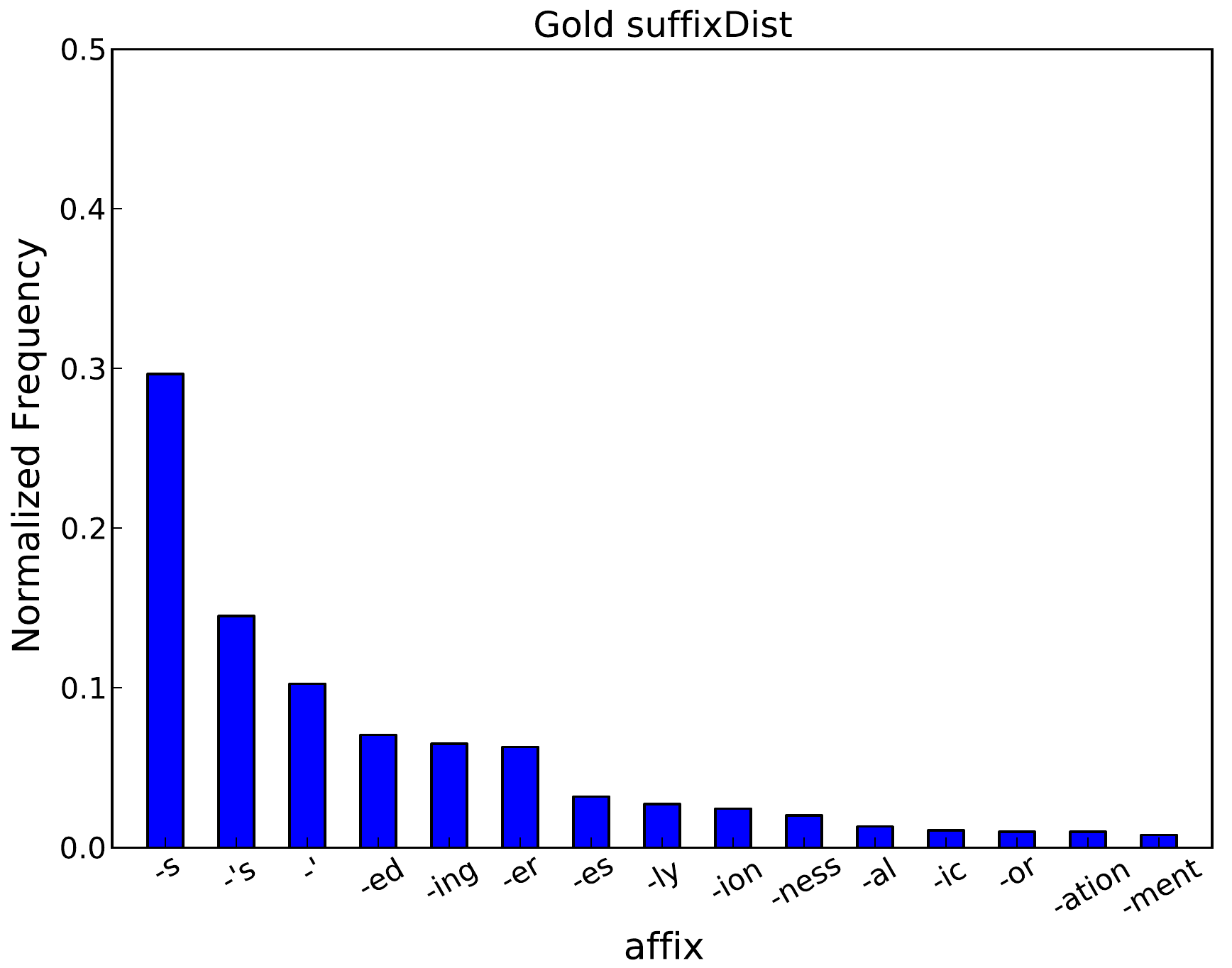}
\label{fig:minipage3}
\end{minipage}
\quad
\begin{minipage}[b]{\linewidth}
\includegraphics[width=\linewidth]{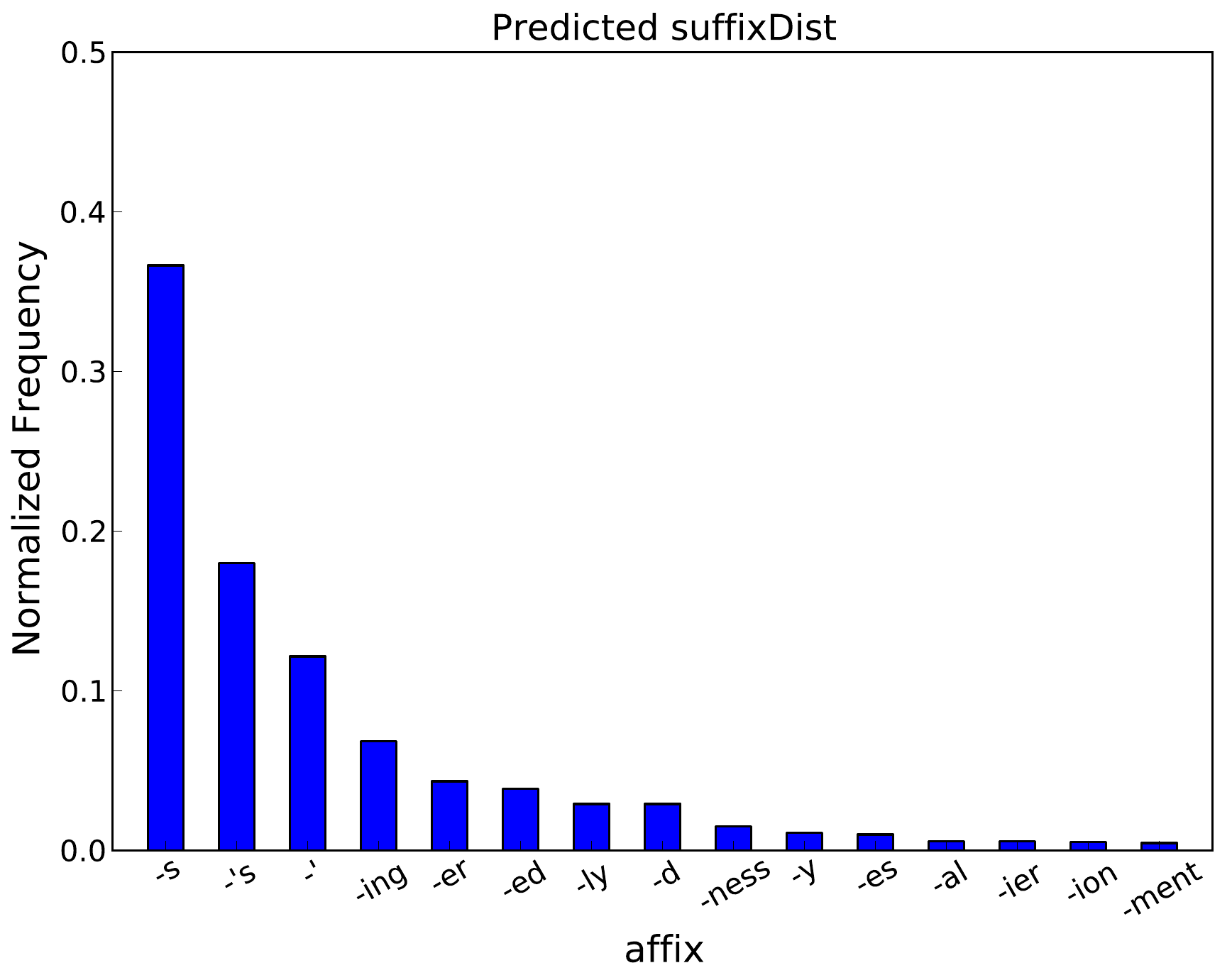}
\label{fig:minipage4}
\end{minipage}
\caption{Comparison of gold and predicted frequency distributions of the top 15 affixes for English}
\label{affix-distribution}
\end{figure}

\paragraph{Affix analysis} We also analyze the various affixes produced by the model, and compare with the gold affixes. Particularly, we plot the frequency distributions of the affixes\footnote{To conserve space, we only show the distribution of suffixes here, but we observe a similar trend for prefixes.} obtained from the gold and predicted segmentations for the English test data in figure~\ref{affix-distribution}. 

 From the figure, we can see that our model learns to identify good affixes for the given language. Several of the top affixes predicted are also present in the gold list, and we also observe similarities in the  frequency distributions.

\section{Conclusion}
In this work, we have proposed a discriminative model for unsupervised morphological segmentation that seamlessly integrates orthographic and semantic properties of words. We use morphological chains to model the word formation process and show how to employ the flexibility of log-linear models to incorporate both morpheme and word-level features, while handling transformations of parent words. Our model consistently equals or outperforms five state-of-the-art systems on Arabic, English and Turkish. Future directions of work include using better neighborhood functions for contrastive estimation, exploring other views of the data that could be incorporated, examining better prediction schemes and employing morphological chains in other applications in NLP.
\section*{Acknowledgements}
We thank Kairit Sirts and Yoong Keok Lee for helping run experiments with their unsupervised morphological analyzers, and Yonatan Belinkov for helping with error analysis in Arabic. We also thank the anonymous TACL  reviewers and members of MIT's NLP group for their insightful comments and suggestions. This work was supported by the Intelligence Advanced
Research Projects Activity (IARPA) via Department of
Defense US Army Research Laboratory contract number
W911NF-12-C-0013. 
The U.S. Government is authorized
to reproduce and distribute reprints for Governmental purposes
notwithstanding any copyright annotation thereon. 
The views and conclusions contained herein are
those of the authors and should not be interpreted as necessarily
representing the official policies or endorsements,
either expressed or implied, of IARPA, DoD/ARL, or the
U.S. Government. 

\bibliography{references}
\bibliographystyle{acl2012}

\end{document}